\documentclass{article}
\usepackage{graphicx}
\usepackage{gb4e}
\usepackage{fullname}
\newcommand{\dopdis}{\textmd{\textsc{dopdis}}}
\newcommand{\card}[1]{\ensuremath{\left|#1\right|}}

\title{Evaluation of the NLP Components of the OVIS2 Spoken\\ Dialogue System}

\author{Gert Veldhuijzen van Zanten$^1$\\
        Gosse Bouma$^2$\\
        Khalil Sima'an$^3$\\
        Gertjan van Noord$^2$\\
        Remko Bonnema$^4$
        }

\date{
$^1$IPO Eindhoven\\
$^2$Rijksuniversiteit Groningen\\
$^3$Universiteit Utrecht\\
$^4$Universiteit van Amsterdam}

\begin{document}
\maketitle
\markboth{\hfill Veldhuijzen, Bouma, Sima'an, van Noord and Bonnema}{Evaluation of the NLP Components in OVIS2}
\begin{abstract}
  The NWO Priority Programme {\em Language and Speech Technology\/} is
  a 5-year research programme aiming at the development of spoken
  language information systems.  In the Programme, two alternative
  natural language processing (NLP) modules are developed in parallel:
  a grammar-based (conventional, rule-based) module and a
  data-oriented (memory-based, stochastic, DOP) module.  In order to
  compare the NLP modules, a formal evaluation has been carried out
  three years after the start of the Programme.  This paper describes
  the evaluation procedure and the evaluation results. The
  grammar-based component performs much better than the data-oriented
  one in this comparison.
\end{abstract}
\section{Introduction}

The NWO Priority Programme {\em Language and Speech Technology\/} is a
5-year research programme aiming at the development of spoken language
information systems. Its immediate goal is to develop a demonstrator
of a public transport information system, which operates over ordinary
telephone lines.  This demonstrator is called OVIS, Openbaar Vervoer
Informatie Systeem ({\em Public Transport Information System\/}). The
language of the system is Dutch.

In this Programme, two alternative NLP modules are developed in
parallel: a grammar-based (conventional, rule-based) module and a
data-oriented (memory-based, stochastic, DOP) module.  Both of these
modules fit into the system architecture of OVIS. They accept as their
input {\em word graphs} produced by the automatic speech recognition
component, and produce {\em updates} which are passed on to the
pragmatic analysis component and dialogue manager.

A word graph (Oerder and Ney, 1993)\nocite{oerder-ney} is a compact
representation for all sequences of words that the speech recogniser
hypothesises for a spoken utterance. The states of the graph represent
points in time, and a transition between two states represents a word
that may have been uttered between the corresponding points in time.
Each transition is associated with an {\em acoustic score}
representing a measure of confidence that the word perceived there was
actually uttered. 

The dialogue manager maintains an information state to keep track of
the information provided by the user. An {\em update} expression is an
instruction for updating the information state. The syntax and
semantics of such updates are defined in Veldhuijzen van Zanten
(1996)\nocite{tst24}.  The sentence:

\begin{exe}
\ex
\gll Ik wil op 4 februari van Amsterdam naar Groningen\\
    I want on 4 February from Amsterdam to Groningen\\
\trans I want to travel from Amsterdam to Groningen on February 4th
\end{exe}

\noindent is translated into the update expression:
\begin{exe}
\ex
\begin{verbatim}
(user.wants.
   (destination.(place.groningen);
   (origin.(place.amsterdam));
   (moment.at.(date.(month.february;day.4)))) 
\end{verbatim}
\end{exe}
\noindent which indicates that the destination and origin slots can be 
filled in, as well as the {\tt moment.at} slot. \\

In order to compare the NLP modules, a formal evaluation has been
carried out three years after the start of the Programme. In this
paper, we first shortly describe the two competing NLP components in
section~\ref{methods}.  The evaluation measures string accuracy,
semantic accuracy and computational resources. This is described in
more detail in section~\ref{procedure}. The evaluation results are
presented in section~\ref{results}. On the basis of these results some
conclusions are drawn in section~\ref{conclusions}.

\section{Two NLP Components}
\label{methods}
For detailed descriptions of the NLP components, the reader is
referred to van Noord et al.\/ (1996a), van Noord et al.\/ (1996b)
\nocite{ovis-deliverable-jan}\nocite{ovis-deliverable-okt}
and van Noord et al.\/ (1999)\nocite{nle} for the grammar-based NLP module. The data-oriented approach is
documented in Scha et al.\/ (1996), Sima'an (1997) and Bod \& Scha (1997).
\nocite{tst31}\nocite{tst35}\nocite{tst38}

\subsection{Data Oriented Parsing}
Research in the Data Oriented Parsing framework explores the
hypothesis that humans analyse new input by drawing analogies with
concrete past language experiences, rather than by applying abstract
rules \cite{Scha}.

In developing computational models embodying this idea, we have so far
focused on one particularly straightforward instantiation of it: our
algorithms analyse new input by considering the various ways in which
this input might be generated by a stochastic process which combines
fragments of trees from an annotated corpus of previously encountered
utterances. Formally, these models may be viewed as implementing
extremely redundant Stochastic Tree Substitution Grammars (STSG's);
the grammar rules are the subtrees extracted from the annotated corpus
\cite{RENS93}.

An important parameter of models of this sort is the way in which the
subtree-substitution probabilities in the stochastic process are
computed on the basis of the subtree frequencies in the annotated
corpus.  All current models follow \cite{RENS93} in defining the
probability of substituting a subtree $t$ on a specific node as the
probability of selecting $t$ among all subtrees in the corpus that
could be substituted on that node, i.e., as the number of occurrences
of $t$ divided by the total number of occurrences of subtrees $t'$
with the same root node label as $t$:
\begin{exe}
\ex\label{bodseq}
$P(t) =\frac{\card{t} }{\sum_{t':\mathit{root}(t')=\mathit{root}(t)}\card{t'}}$
\end{exe}

Given these subtree substitution probabilities, the probability of a
derivation $t_1 \circ\cdots\circ t_n$ can be computed as the product
of the probabilities of the substitutions that it consists of
\begin{exe}
\ex
$P(t_1\circ\cdots\circ t_n) = \prod_i P(t_i)$
\end{exe}
\noindent The probability of a
parse tree is equal to the probability that any of its distinct
derivations is generated, which is the sum of the probabilities of all
derivations of that parse tree.  Let $t_{\mathit{id}}$ be the $i$-th
subtree in the derivation $d$ that yields tree $T$, then the
probability of $T$ is given by: 
\begin{exe}
\ex
$P(T) = \sum_d \prod_i P(t_{\mathit{id}})$
\end{exe}

An efficient polynomial algorithm for computing the Most Probable
Derivation is given in Sima'an (1996a).\nocite{Simaan5} 
From a theoretical point of
view we might expect the computation of the Most Probable Parse to
yield better disambiguation accuracies than the Most Probable
Derivation, and this expectation was confirmed by certain experiments.
However, Sima'an (1996b)\nocite{Simaan4} has shown that the 
problem of computing the
Most Probable Parse is not solvable by deterministic polynomial time
algorithms.  For reasons regarding time-complexity, the most probable
derivation (MPD) is still the method of choice for a real-world
application.

The algorithm presented in Sima'an (1996a)\nocite{Simaan5} is
implemented in the ``Data Oriented Parsing and DIsambiguation System''
(\dopdis).  The algorithm extends a well-known CFG parsing algorithm
(the CKY algorithm) in a suitable way in order to deal with STSG's.
The extension makes use of the fact that the paths in a parse-tree,
which is generated from an STSG derivation for a given sentence, form
a regular set that can be easily computed.  By computing such sets,
\dopdis\ generates exactly the necessary trees which the STSG
dictates. On top of this mechanism, \dopdis\ computes both the most
probable derivation and the probability of an input sentence.  The
construction operates in time-complexity which is cubic in sentence
length and linear in STSG size, which is a good achievement in parsing
tree grammars (existing tree-parsing techniques have complexity which
is square in grammar size).

The extension to parsing and disambiguating word graphs maintains the
same time and space complexity (where instead of sentence length here
the complexity concerns the numbers of nodes i.e.\ states the
word graph contains).  \dopdis\ computes the so called Most Probable
intersection-Derivation of a word graph and the DOP STSG. An
intersection-derivation (i-derivation) is a pair
\mbox{$\langle\mathit{string, derivation}\rangle$} 
where $\mathit{string}$ is the
string of words on a path in the input word graph, and
$\mathit{derivation}$ is an STSG derivation for that string. The
probability of the i-derivation is computed through a weighted product
of the probabilities on the path in the word graph and the
STSG-derivation probability. The probabilities of the word graph paths
are obtained from the speech-recognisers likelihoods by a
normalisation heuristic. The probabilities resulting from this
heuristic combine better with the DOP probabilities than raw speech
recogniser likelihoods. The issue of scaling the likelihoods of the
speech-recogniser in a well-founded way is still under study. The
current method divides the likelihood of every transition by a
normalisation factor that is computed from the word graph.

\paragraph{An extension to semantic interpretation.}
In van den Berg, Bod and Scha (1994)\nocite{Berg94}, an extension of
the model to semantic interpretation is presented.  A first
implementation of this extension was described in Bonnema
(1996)\nocite{bonnema:thesis}.  A DOP model as described above can be
extended from just syntactic, to semantic analysis, by augmenting the
trees in the tree-bank with semantic \emph{rules}. These rules
indicate, for each individual analysis in the tree-bank, how its
meaning is constructed out of the meaning of its parts.  Just as in
the purely syntactic version of DOP, we extract all possible fragments
from these syntactic/semantic analyses. We then use these fragments to
build an STSG. We alter the constraints on tree substitution, by
demanding that both the syntactic category and the semantic type of
the root node of a subtree match with those at the substitution site.
Note that the semantic types restrict the possibility of substitution.
The language generated by an STSG created on the basis of a
semantically enriched tree-bank, is therefore a subset of the language
generated by an STSG created on the basis of the same tree-bank
without the semantic annotations. Ideally, the former STSG would
exclude exactly the set of semantically ill-formed sentences.  The
preferred analysis of an utterance will now provide us with both a
syntactic and a semantic interpretation.  In the current
implementation the most probable analysis is taken to be the
interpretation given by the most probable derivation.

A corpus of syntactic and semantic analyses of transcribed
utterances from the OVIS domain was created to test this model.  The
OVIS tree-bank currently contains 10.000 analyzed utterances.  The
top-node semantics of each annotated utterance is an
\emph{update-expression} that conforms to the formalism described
in Veldhuijzen van Zanten (1996).  The semantic label of a node $N$ in an analysis
consists of a rule, that indicates how the meaning of $N$ (the
\emph{update}) is built-up out of the meanings of $N$'s daughter
nodes.  This semantic rule is typed. Its type follows from both the
rule itself, and from the types of the semantic labels of the
daughter-nodes of $N$, given the definition of the logical language
used.  In the present case, the type of an expression is a pair of
integers, its \emph{meet} and \emph{join}. The \emph{meet} and
\emph{join} correspond to the \emph{least upper bound} and the
\emph{greatest lower bound} of the expression in the type-hierarchy,
as described in Veldhuijzen van Zanten (1996).

A semantically enriched STSG as described above, must fulfil an
important property.  It has to be possible to define a function from
derivations to logical formulas, that is defined for \emph{every}
derivation that can be produced by the grammar.  In other words, the
information provided by semantic types and syntactic categories in an
analysis must be sufficient.  Because the set of subtrees is closed
under the operation of subtree extraction, i.e., all subtrees~$T'$
that can be extracted from another subtree~$T$, belong to the same set
as $T$, it is easy to establish this property, even for a very large
grammar.  We only need to look at the subtrees of depth one.  If there
is a unique semantic rule associated with the root-node of all
subtrees of depth one, given the syntactic categories and semantic
types of its nodes, it follows that we know the semantic rule at the
nonterminal-nodes of every subtree. Fortunately, the nature of the
annotated tree-bank is such, that in about 99.9~\% of cases we can
indeed establish the semantic rule at the root-node of a subtree in
this way. The few exceptions are assigned an ``exception-type'', to
reduce the uncertainty to zero.  We exploit the property described
above, to construct a rewrite system for the semantic STSG.  This
rewrite system applies the semantic rules associated with every node
in a derivation in a bottom up fashion, and arrives at the complete
logical formula.

\paragraph{Methods for word graphs.}
The evaluation experiments were performed using just the semantic
DOP-model as it was described above.

For every word graph the most probable intersection derivation was
determined.  The leaf-nodes of this derivation constitute the best
path through the word graph.  The probability of a derivation is
calculated on the basis of both the probabilities of the subtrees
extracted from the OVIS tree-bank, and the acoustic likelihoods of the
words in the word graph.

We created several instances of the semantic DOP-parser, with
differing constraints on the form of possible subtrees.  Four
parameters can be distinguished, whose values determine the
constraints on subtrees.  Below the parameters are given, with the
letters we commonly use to refer to each parameter.
\begin{description}
\item{{\bf d}} The maximal depth of subtrees.
\item{{\bf l}} The maximal number of lexical items in a subtree.
\item{{\bf L}} The maximal number of consecutive lexical items in a subtree.
\item{{\bf n}} The maximal number of substitution sites in a subtree. 
\end{description}

Obviously, the number of possible combinations is huge.  We chose to
use the parameter settings for which previous experiments yielded the
best results.

The results presented in this document are all obtained using the
following settings: {\bf l}=9, {\bf L}=3, {\bf n}=2.  For the maximum
depth we used {\bf d}=2 and {\bf d}=4. No constraints were applied to
subtrees of depth 1.

If a word graph contains more than 350 transitions, the maximum depth
of subtrees is automatically limited to \emph{two}, to avoid excessive
memory requirements. About 3\% of word graphs in the testing material
do contain more than 350 transitions.

\paragraph{Methods for test sentences.}
For sentences, we used the same parameter settings, but added an
experiment with {\bf d}=5.

In this document, some results on sentences are given with the extra
indication \emph{group}.  We will now briefly explain what this
means.

Non-terminals in the semantic DOP-model consist of a
syntactic-category / semantic-type pair.  Such a non-terminal imposes
a rather rigid constraint on substitution. For the parsing of
word graphs, this constraint seems to be beneficiary.  For the parsing
of sentences, on the other hand, these constraints could be too rigid.
A greater degree of freedom results in over-generation, which in turn
may lead to better statistics.  An algorithm was designed to group
semantic types that have a comparable distribution.  This results in
fewer non-terminals in the tree-bank, and has been shown to lead to a
higher semantic accuracy for sentences.  The results marked with
\emph{group} indicate that this grouping algorithm has been employed.

\subsection{Grammar-based NLP}

The grammar-based NLP component developed in Groningen is based on a
detailed computational grammar for Dutch, and a robust parsing
algorithm which incorporates this grammatical knowledge as well as
other knowledge sources, such as the acoustic evidence (present in the
word graph) and Ngram statistics (collected from a large set of user
utterances). It has been argued (van Noord et al.\/ 1999)\nocite{nle}
that robust parsing can be based on sophisticated grammatical
analysis. In particular, the grammar describes full sentences, but in
doing so, also describes the grammar of temporal expressions and
locative phrases which are the crucial concepts for the timetable
information application. Robustness is achieved by taking these
phrases into consideration, if a full parse of an utterance is not
available.

\paragraph{Computational Grammar for Dutch.}

In developing the  grammar the
short-term goal of developing a grammar which meets the requirements
imposed by the application (i.e.\ robust processing of the output of
the speech recogniser, extensive coverage of locative phrases and
temporal expressions, and the construction of fine-grained semantic
representations) was combined with the long-term goal of developing a general,
computational, grammar which covers all the major constructions of
Dutch.

The design and organisation of the grammar, as well as many aspects of
the particular grammatical analyses, are based on
Head-driven Phrase Structure Grammar (Pollard and Sag
1994)\nocite{hpsg2}. The grammar is compiled
into a restricted kind of definite clause grammar for which efficient
processing is feasible. The semantic component follows the approach to
monotonic semantic interpretation using {\em quasi-logical forms}
presented originally in Alshawi (1992)\nocite{cle-book}.

The grammar currently covers the majority of verbal subcategorisation
types (intransitives, transitives, verbs selecting a {\sc pp}, and
modal and auxiliary verbs), {\sc np}-syntax (including pre- and
post-nominal modification, with the exception of relative clauses),
{\sc pp}-syntax, the distribution of {\sc vp}-modifiers, various
clausal types (declaratives, yes/no and {\sc wh}-questions, and
subordinate clauses), all temporal expressions and locative phrases
relevant to the domain, and various typical spoken-language
constructs.  Due to restrictions imposed by the speech recogniser, the
lexicon is relatively small (3200 word forms, many of which are names
of stations and cities). 

\paragraph{Robust and Efficient Parsing.}

Parsing algorithms for strings can be generalised to parse 
word graphs (van Noord 1995)\nocite{acl95}.  In
the ideal case, the parser will find a path in the word graph that can
be assigned an analysis according to the grammar, such that the path
covers the complete time span of the utterance, i.e.\ the path leads
from the start state to a final state. The analysis gives rise to an
update of the dialogue state, which is then passed on to the dialogue
manager.

However, often no such paths can be found in the word graph,
due to:
\begin{itemize}
\item errors made by the speech recogniser,
\item linguistic constructions not covered in the grammar, and
\item irregularities in the spoken utterance.
\end{itemize}

Even if no full analysis of the word graph is possible, it is usually 
the case that useful information can be extracted from the word graph.
Consider for example the utterance: 

\begin{exe}
\ex
\gll Ik wil van van Assen naar Amsterdam\\
I want from from Assen to Amsterdam\\
\trans I want to travel from Assen to Amsterdam
\end{exe}

The grammar will not assign an analysis to this utterance due to the
repeated preposition. However, it would be useful if the parser would
discover the prepositional phrases {\em van Assen} and {\em naar
  Amsterdam} since in that case the important information contained in
the utterance can still be recovered. Thus, in cases where no full
analysis is possible the system should fall back on an approach
reminiscent of concept spotting. In van Noord et al.\/ (1999) \nocite{nle} a
general algorithm is proposed which achieves this.

The first ingredient to a solution is that the parser is required to
discover all occurrences of major syntactic categories (such as noun
phrase, prepositional phrase, subordinate sentence, root sentence)
{\em anywhere in the word graph}. Conceptually, one can think of these
categories as edges which are added to the word graph in addition to
the transitions produced by the speech recogniser.

For such word graphs annotated with additional category edges, a path
can be defined as a sequence of steps where each step is either a
transition or a category edge. A transition step is called a `skip'. 
For a given annotated word graph many
paths are possible. On the basis of an appropriate {\em weight}
function on such paths, it is possible to search for the {\em best}
path. The search algorithm is a straightforward
generalisation of the {\sc dag-shortest-path} algorithm (Cormen et al.\/ 1990). \nocite{algorithms}

The weight function is sensitive to the following factors:
\begin{itemize}
\item Acoustic score. Obviously, the acoustic score 
  present in the word graph is an important factor. 
\item The number of skips is minimised
  in order to obtain a preference for the maximal projections found by
  the parser. 
\item Number of maximal projections. The number of maximal projections
  is minimised in order to obtain a preference for more extended
  linguistic analyses over a series of smaller ones.
\item Ngram statistics. 
\end{itemize}

The grammar-based NLP component is implemented in SICStus Prolog.
Below we report on a number of different methods which are all
variations with respect to this weight function.

\paragraph{Variants of Grammar-based NLP.}

The grammar-based NLP methods that have taken part in the evaluation
are of two types. The first type, {\em b(B,N)}, consists of two
phases. In the first phase the word graph is made smaller by selecting
the N-best paths from the word graph, using the acoustic scores and a
language model consisting of bigrams (B=bi) or trigrams (B=tri) (with
bigrams for backing-off).  Only those transitions of the word graph
remain which are part of at least one of those N-best paths. In the second
phase the parser is applied, using acoustic scores and a language
model of trigrams (again with bigrams for backing-off).

The second type of method is {\em f(B,N)}. In this case, if the
word graph contains less than N transitions, then the full word graph
is input to the parser, and acoustic scores and a language model of
trigrams (bigrams for backing-off) is applied to select the best
analysis. If the word graph contains more than N transitions, then
method b(B,1) is applied. \\

\section{Evaluation Procedure and Criteria}
\label{procedure}

\subsection{Procedure}

An experimental version of the system has been available to the
general public for almost a year. From a large set of more recent
dialogues a subset was selected randomly for testing. Many of the
other dialogues were available for training purposes. Both the
training and test dialogues are therefore dialogues with `normal'
users. 

In particular, a training set of 10K richly annotated word graphs was
available.  The 10K training corpus is annotated with the user
utterance, a syntactic tree and an update.  This training set was used
to train the DOP system. It was also used by the grammar-based
component for reasons of grammar maintenance and grammar testing.

A further training set of about 90K annotated user utterances was
available as well. It was primarily used for constructing the Ngram
models incorporated in the grammar-based component.

The NLP components of OVIS2 have been evaluated on 1000 unseen user
utterances. The latest version of the speech recogniser produced 1000
word graphs on the basis of these 1000 user utterances.  For these
word graphs, annotations consisting of the actual sentence ('test
sentence'), and an update ('test update') were assigned
semi-automatically, without taking into account the dialogue context
in which the sentences were uttered.  These annotations were unknown
to both NLP groups. The annotation tools are described in
Bonnema (1996).\nocite{bonnema:thesis}

After both NLP components had produced the results on word graphs, the 
test sentences were made available. Both NLP components were then
applied to these test utterances as well, to mimic a situation in
which speech recognition is perfect. 

The test updates were available for inspection by the NLP groups only
after both modules completed processing the test material. A small
number of errors was encountered in these test updates. These errors
were corrected before the accuracy scores were computed.  The accuracy
scores presented below were all obtained using the same evaluation
software.

\subsection{Criteria}

The NLP components were compared with respect to the following two
tasks. Note that in each task, analysis proceeds in isolation from the
dialogue context. The first task is to provide an update for the test
sentence (in this report we refer to this update as the `best
update'). The second task is to provide an update and a sentence for
the word graph (`best update' and `best sentence'). The quality of the
NLP components will be expressed in terms of string accuracy
(comparison of the best sentences with the test sentences), semantic
accuracy (comparison of the best updates with the test updates) and
computational resources. Each of these criteria is now explained in
more detail.

\paragraph{String accuracy.} 
String accuracy measures the distance of the test sentence and the
best sentence. String accuracy is expressed in terms of sentence
accuracy (SA, the proportion of cases in which the test sentence
coincides with the best sentence), and word accuracy (WA).  The string
comparison on which word accuracy is based is defined by the minimal
number of substitutions, deletions and insertions of words that is
required to turn the best sentence into the test sentence (Levenshtein
distance). Word accuracy is defined as

\begin{exe}
\ex $\mbox{\it WA} = 1 - \frac{d}{n}$
\end{exe}
\noindent where $n$ is the length of the actual utterance and $d$ is the
Levenshtein distance. For example, if the analysis gives 'a b a c d'
for the utterance 'a a c e', then the Levenshtein distance is 2, hence
the WA is 1-2/4 is 50\%.

\paragraph{Semantic accuracy.}
An update is a logical formula which can be evaluated against an
information state and which gives rise to a new, updated information
state. The most straightforward method for evaluating concept accuracy
in this setting is to compare the update produced by the grammar with
the annotated update. One problem with this approach is the fact that
the update language does not provide a simple way to compute
equivalence of updates (there is no notion of normal form for update
expressions). A further obstacle is the
fact that very fine-grained semantic distinctions can be made in the
update-language.  While these distinctions are relevant semantically
(i.e.\ in certain cases they may lead to slightly different updates of
an information state), they often can be ignored by a dialogue
manager. For instance, the updates below are semantically not
equivalent, as the ground-focus distinction is slightly different. In
the first update the feature {\tt place} is supposed to be {\em
  ground}, whereas in the second update, it is part of the {\em
  focus}. 

\begin{exe}
\ex
\begin{verbatim}
user.wants.travel.destination.place
   ([# town.leiden];[! town.abcoude])
\end{verbatim}
\ex
\begin{verbatim}
user.wants.travel.destination.
   ([# place.town.leiden];[! place.town.abcoude])
\end{verbatim}
\end{exe}
However, the dialogue manager will decide in both cases that
this is a correction of the destination town. 

Since semantic analysis is the input for the dialogue manager, we have
therefore measured concept accuracy in terms of a simplified version
of the update language. Following a somewhat similar proposal in Boros
et al.\/ (1996)\nocite{Boros}, we translate each update into a set of
``semantic units'', were a unit in our case is a triple
$\langle\mbox{\it CommunicativeFunction Slot Value}\rangle$.  For instance, the examples
above translate as
\begin{exe}
\ex
\begin{verbatim}
<denial destination_town leiden>
<correction destination_town abcoude>
\end{verbatim}
\end{exe}
Both the updates in the annotated corpus and the updates produced by
the system are translated into semantic units of the form given above.
The syntax of the semantic unit language and the translation of
updates to semantic units is defined in van Noord
(1997)\nocite{tst46}, but note that the translation of updates to
semantic units is relatively straightforward and is not expected to be
a source of discussion, because the relation is many to one.

Semantic accuracy can now be defined as follows.  Firstly, we list
the proportion of utterances for which the corresponding semantic
units exactly match the semantic units of the annotation ({\em exact
  match}). Furthermore we calculate {\em precision} (the number of
correct semantic units divided by the number of semantic units which
were produced) and {\em recall} (the number of correct semantic units
divided by the number of semantic units of the annotation).  Finally,
following Boros et al.\/ (1996)\nocite{Boros} we also present concept accuracy as

\begin{exe}
\ex
$\mbox{\it CA} = \left( 1 - \frac{SU_S + SU_I + SU_D}{SU} \right)$
\end{exe} 

\noindent where $SU$ is the total number of semantic units in the corpus
annotation, and $SU_S$, $SU_I$, and $SU_D$ are the number of
substitutions, insertions, and deletions that are necessary to make
the (translated) update of the analysis equivalent to (the translation
of) the corpus update. 

\paragraph{Computational Resources.} 
In order to measure computational efficiency, the total amount of
CPU-time, the maximum amount of CPU-time per input, and the total memory
requirements will be measured. Due to differences in hardware, details 
differ between the two NLP components. 

For the data-oriented methods, the CPU-time given is the \emph{user-time}
of the parsing process, in seconds.  This measure excludes the time
used for system calls made on behalf of the process (this can be
ignored).  Time was measured on a Silicon Graphics \emph{Indigo}, with
a MIPS R10000 processor, running IRIX 6.2.  Memory usage is the
maximum number of mega-bytes required, to interpret the 1000
utterances.  Regrettably, for a very small percentage (0.02\%) of
word graphs, the process ran out of memory. This means that the figures
for word graph parsing indicate the size of the jobs at the moment the
system gave up, which is generally when the physical memory is filled.
On the other hand, we should acknowledge the fact that some large
word graphs that \emph{did} receive an interpretation, also approached
this limit.

For the grammar-based  methods, CPU-time is measured in milliseconds on a
HP 9000/780 (running HP-UX 10.20). The system uses SICStus Prolog 3
\#3. CPU-time include all phases of processing, but does not contain
the time required for system calls (can be ignored) and garbage
collection (adds at most 15\% for a given run). The memory
requirements are given as the increase of the UNIX process size to
complete the full run of 1000 inputs. At start-up the process size can
be as large as 30 megabytes, so this number has been added in order to
estimate total memory requirements.

\begin{table}
\begin{center}
\begin{tabular}{|c|r|r|r|r|r|r|r|}\hline
& graphs & trans & states & words & t/w & max(t)& max(s)\\\hline
input 
& 1000  &  48215       & 16181  & 3229 & 14.9 & 793 & 151 
\\
normalised
& 1000  &  73502       & 11056  & 3229 & 22.8 & 2943 & 128\\     
\hline
\end{tabular}
\end{center}
\caption{\label{een}
{\bf Characterisation of test set (1).}
This table lists the number of transitions, the number of states, 
the number of words of the actual utterances, the average number of
transitions per word, the maximum number of transitions, and the 
maximum number of states.  The first row provides those statistics 
for the input word graph; the second row 
for the so-called normalised word graph in which all 
$\epsilon$-transitions (to model the absence of sound) are removed.  
The number of transitions per word is an indication of the extra 
ambiguity for the parser  introduced by the word graphs in comparison 
with parsing of an ordinary string. }
\end{table}

\subsection{Test Set}
Some indication of the difficulty of the set of 1000 word graphs is
presented in table~\ref{een}.
A further indication of the difficulty of this set of word graphs is
obtained if we look at the word and sentence accuracy obtained by a
number of simple methods.  The method {\em speech} only takes into
account the acoustic scores found in the word graph. No language model
is taken into account. The method {\em possible} assumes that there is
an oracle which chooses a path such that it turns out to be the best
possible path.  This method can be seen as a natural upper bound of
what can be achieved.

The methods {\em speech\_bigram} and {\em speech\_trigram} use a
combination of bigram (resp. trigram) statistics and the speech score.
In the latter four cases, a language model was computed from about
50K utterances (not containing the utterances from the test set).
The results are summarised in table~\ref{two}. 
\begin{table}
\small
\begin{center}
\begin{tabular}{|c|r|r|}\hline
method               &        WA &      SA\\\hline
speech               &      69.8 &  56.0\\
possible             &      90.5 &  83.7\\
speech\_bigram       &      81.1 &  73.6\\
speech\_trigram      &      83.9 &  76.2\\\hline
\end{tabular}
\end{center}
\caption{\label{two}{\bf Characterisation of test set (2).}
Word accuracy and sentence accuracy based on
  acoustic score only (speech); using the best
  possible path through the word graph, i.e.\/ based on acoustic scores only
  (possible); and
  using a combination of bigram (resp. trigram) scores and acoustic
  scores. }
\end{table}

During the development of the NLP components of OVIS2, word graphs
were typically small: about 4 transitions per word on average. During
the evaluation, however, the number of transitions per word for the
test set was much larger. It turned out that the NLP components had
trouble with very large word graphs (both memory and CPU-time
requirements increase rapidly). 

Recently, improvements have already been obtained to treat such large
word graphs.  For example, the grammar-based NLP component has been
extended with a heuristic version of the search algorithm which is {\em not}
guaranteed to find the best path. In practice this implementation
returns the same answers as the original search algorithm, but much
more quickly so (two orders of magnitude faster).

\section{Results of the Evaluation}
\label{results}

This section lists the results for word graphs. In table~\ref{wgwa} we 
list the results in terms of string accuracy, 
semantic accuracy 
and the computational resources required to complete the test.

\begin{table}
\footnotesize
\begin{center}
\begin{tabular}{|c|c|r|r|r|r|r|r|r|r|r|}\hline
Method             &     Site  & \multicolumn{2}{c|}{String Acc}&
                                 \multicolumn{4}{c|}{Semantic Accuracy}&
                                 \multicolumn{2}{c|}{CPU} &
                                 Mem\\
             &      &     WA &   SA & match & prec &
             recall & ca & total & max & max\\
\hline
d2          & A'dam & 76.8 & 69.3& 74.9 & 80.1 & 78.8 & 75.5&   7011 &  648 &  619\\
d4          & A'dam & 77.2 & 69.4& 74.9 & 79.1 & 78.8 & 75.1&  32798 & 2023 &  621\\
f(bi,50)        & Gron & 81.3 & 74.6& 79.5 & 82.9 & 83.8 & 79.9&    215 &   16 &     37\\
f(bi,100)       & Gron & 82.3 & 75.8& 80.9 & 83.6 & 84.8 & 80.9&    297 &   15 &     37\\
f(bi,125)       & Gron & 82.3 & 75.9& 81.3 & 83.9 & 85.2 & 81.3&    340 &   24 &     38\\
b(bi,1)         & Gron & 81.1 & 73.6& 78.5 & 82.1 & 83.1 & 78.9&    175 &   16 &     31\\
b(bi,2)         & Gron & 82.3 & 75.7& 80.8 & 83.9 & 84.8 & 81.1&    255 &   20 &     32\\
b(bi,4)         & Gron & 82.8 & 76.0& 80.8 & 83.8 & 85.0 & 81.3&    479 &  115 &     34\\
b(bi,8)         & Gron & 83.4 & 76.5& 81.6 & 84.6 & 85.6 & 82.2&    780 &  276 &     43\\
b(bi,16)        & Gron & 83.8 & 76.4& 81.7 & 84.9 & 86.0 & 82.6&   1659 &  757 &     60\\
f(tr,50)       & Gron & 83.9 & 76.2& 81.8 & 84.9 & 85.9 & 82.5&   1399 &  607 &     64\\
f(tr,100)      & Gron & 84.2 & 76.6& 82.0 & 85.0 & 86.0 & 82.6&   1614 &  690 &     64\\
f(tr,125)      & Gron & 84.2 & 76.5& 82.1 & 85.3 & 86.3 & 82.8&   1723 &  755 &     64\\
b(tr,1)        & Gron & 83.9 & 76.2& 81.5 & 84.5 & 85.7 & 82.2&   1420 &  603 &     64\\
b(tr,2)        & Gron & 84.1 & 76.4& 81.8 & 85.3 & 86.4 & 83.0&   2802 & 1405 &    101\\
b(tr,4)        & Gron & 84.3 & 76.4& 82.0 & 85.4 & 86.4 & 83.0&   5524 & 2791 &    177\\
\hline
\end{tabular}
\end{center}
\caption{\label{wgwa}{\bf Accuracy and Computational Resources for 1000 word graphs.}
String Accuracy and Semantic Accuracy is given as percentages;  total
and maximum CPU-time in seconds, maximum memory requirements in 
Megabytes.}
\end{table}

The total amount of CPU-time is somewhat misleading because typically
many word graphs can be treated very efficiently, whereas only a few
word graphs require very much CPU-time.
In table~\ref{wgcom} we
indicate the semantic accuracy (concept accuracy) that 
is obtained if a time-out is assumed (in such 
cases we assume that the system does not provide an update).

\begin{table}
\small
\begin{center}
\begin{tabular}{|c|c|r|r|r|r|r|r|}\hline
Method        &     Site  &  100 &  500 & 1000 & 5000 & 10000 & $>$  \\\hline
d2    &A'dam & 37.0& 53.0& 58.1& 68.1& 70.4 & 75.5 \\
d4    &A'dam & 24.6& 34.5& 38.2& 50.4& 57.3 & 75.1 \\
f(bi,50)   & Gron & 46.0& 73.7& 76.9& 80.3& 80.3 & 79.9 \\
f(bi,100)  & Gron & 44.4& 67.9& 75.3& 81.1& 81.2 & 80.9 \\
f(bi,125)  & Gron & 44.6& 64.9& 73.3& 81.3& 81.7 & 81.3 \\
b(bi,1)    & Gron & 58.2& 73.1& 76.6& 79.3& 79.2 & 78.9 \\
b(bi,2)    & Gron & 54.7& 74.1& 77.6& 81.1& 81.5 & 81.1 \\ 
b(bi,4)    & Gron & 49.6& 72.3& 75.6& 80.3& 80.5 & 81.3 \\
b(bi,8)    & Gron & 45.9& 70.2& 74.4& 80.9& 81.5 & 82.2 \\
b(bi,16)   & Gron & 42.2& 65.5& 72.5& 78.0& 81.0 & 82.6 \\
f(tr,50)  & Gron & 45.5& 71.2& 75.4& 81.0& 81.7 & 82.6 \\
f(tr,100) & Gron & 44.5& 64.2& 71.9& 80.5& 81.8 & 82.6 \\
f(tr,125) & Gron & 44.1& 62.2& 70.2& 80.6& 81.9 & 82.8 \\
b(tr,1)   & Gron & 52.7& 70.9& 74.7& 80.7& 81.2 & 82.2 \\
b(tr,2)   & Gron & 49.6& 68.8& 72.7& 79.1& 81.4 & 83.0 \\
b(tr,4)   & Gron & 48.0& 66.6& 71.6& 78.2& 79.5 & 83.0 \\
\hline
\end{tabular}
\end{center}
\caption{\label{wgcom}{\bf Concept accuracy for 1000 word graphs} (percentages), if all results are disregarded with a time-out of respectively 100,
  500, 1000, 5000, 10000 milliseconds of CPU-time. The last column
  repeats the results if no time-out is assumed.}
\end{table}


We also present the results for test sentences (rather than
word graphs).  Such a test indicates what the results are if the
speech recogniser would perform perfectly.  Obviously, it does not
make sense to measure string accuracy in such a set-up. Semantic
accuracy and computational resources is presented in
table~\ref{sentca}. Because the average sentence length is very small, 
we present the results for
concept accuracy versus the length of the input sentence in
table~\ref{sentcan}. 


\begin{table}
\small
\begin{center}
\begin{tabular}{|c|c|r|r|r|r|r|r|r|}\hline
Method   &     Site  &  \multicolumn{4}{c|}{Semantic Accuracy}&
                                 \multicolumn{2}{c|}{CPU} &
                                 Mem\\
         &       &   match & prec & recall & ca & total & max & max\\\hline
d4       & A'dam &   92.2 &  93.8 &  91.2 & 90.4 & 856 & 14 & 21\\
group.d2 & A'dam &   93.0 &  94.0 &  92.5 & 91.6 &  91 &  9 & 14\\
group.d4 & A'dam &   92.7 &  93.8 &  91.8 & 91.0 &1614 &174 & 48\\
group.d5 & A'dam &   92.6 &  93.7 &  92.3 & 91.4 &3159 &337 & 78\\
nlp      &  Gron &   95.7 &  95.7 &  96.4 & 95.0 &  27 &  1 & 31\\
\hline
\end{tabular}
\end{center}
\caption{\label{sentca}{\bf Semantic Accuracy and Computational 
Resources for 1000 test sentences.} Total and maximum CPU-time in seconds; 
memory in Megabytes.}
\end{table}

\begin{table}
\small
\begin{center}
\begin{tabular}{|c|c|r|r|r|r|r|r|}
\hline
Method & site & all & $\geq 2$ & $\geq 4$ & $\geq 6$ & $\geq 8$ & $\geq 10$\\ 
\hline
\multicolumn{2}{|c|}{\# instances} 
                   & 1000 &  601 &  344 &  160 &  74 &    38\\
\hline
d4       & A'dam   & 90.4 & 87.4 & 84.7 & 75.9 & 69.8 & 65.2\\
group.d2 & A'dam   & 91.6 & 89.0 & 86.7 & 78.7 & 69.8 & 68.3\\
group.d4 & A'dam   & 91.0 & 88.2 & 85.8 & 77.3 & 69.4 & 64.6\\
group.d5 & A'dam   & 91.4 & 88.7 & 86.6 & 78.9 & 71.8 & 67.1\\
     nlp & Gron    & 95.0 & 93.4 & 93.0 & 88.1 & 85.9 & 87.0\\
\hline
\end{tabular}
\end{center}
\caption{\label{sentcan}{\bf Concept Accuracy versus Sentence Length
    for 1000 test sentences.} The third column repeats the results for 
  the full test set. The remaining columns list the results for the
  subset of the test set containing the sentences with at least 2
  (4, 6, 8, 10) words.}
\end{table}



\section{Conclusions}
\label{conclusions}

The grammar-based methods developed in Groningen perform much better
than the data-oriented methods developed in Amsterdam. For word
graphs, the best data-oriented method obtains an error-rate for
concept accuracy of 24.5\%. The best grammar-based method performs
more than 30\% better: an error-rate for concept accuracy of 17.0\%.
For sentences, a similar difference can be observed. The best
data-oriented method obtains an error rate for concept accuracy of
8.5\% whereas the grammar-based method performs more than 40\% better
with a 5.0\% error rate. The differences increase with increasing
sentence length.

The grammar-based methods require less computational
resources than the data-oriented methods.  However, 
the CPU-time requirements are still outrageous for a small
number of very large word graphs\footnote{As mentioned before, a 
  dramatic reduction has been obtained by a heuristic search algorithm.}. 
For sentences, the grammar-based component
performs satisfactorily (with a maximum CPU-time of 610
milliseconds).
  
The by far most important problem for the application consists of
disambiguation of the word graph. The evaluation shows that NLP hardly
helps here: a combination of speech scores and trigram scores performs
much better in terms of string accuracy than the data-oriented
methods. 
The grammar-based methods have incorporated the
insight that Ngrams are good at disambiguating word graphs; by
incorporating Ngram statistics similar results for string accuracy are
obtained. In order to see whether NLP helps at all, we could compare
the b(tr,1) method (which simply uses the best path in the
word graph as input for the parser) with any of the other
grammar-based methods. For instance, the method b(tr,4) performs
somewhat better than b(tr,1) (83.0\% vs.\/  82.2\% concept
accuracy). This shows that in fact NLP is helpful in choosing the best
path\footnote{This result is (just) statistically significant. We
  performed a paired T-test on the number of wrong semantic units per
  graph. This results in a $t^*$ score of 2.0 (with 999 degrees of
  freedom).}.  If it were feasible to use methods b(tr,N) or
f(tr,N) with larger values of N, further improvements might be
possible.
  
Once a given word graph has been disambiguated, then both NLP
components work reasonably well: this can be concluded based upon the
concept accuracy obtained for sentences. In those cases the
grammar-based NLP component also performs better than the data-oriented
parser; this indicates that the difference in
performance between the two components is not (only) due to the
introduction of Ngram statistics in the grammar-based NLP component.

The current evaluation has brought some important shortcomings of the
DOP approach to light. Two important problems, for which solutions are
in the making, are briefly discussed below.

The first one is the inadequacy of the definition of subtree
probability.  It turns out that Bod's equation~(\ref{bodseq}) given 
on page~\pageref{bodseq} shows a bias toward analyses derived by
subtrees from large corpus trees.  The error lies in viewing an
annotated corpus as the ``flat'' collection of all its subtrees.
Information is lost when the distribution of the analyses that supply
the subtrees is ignored. The effect is that a large part of the
probability mass is consumed by subtrees stemming from relatively
rare, large trees in the tree-bank.  A better model has been designed,
that provides a more reliable way of estimating subtree probabilities.

The second shortcoming we will discuss is 
the fact that existing DOP algorithms are unable to generalise over the     
syntactic structures in the data.
Corpus-based methods
such as the current implementation of DOP, assume that the tree-bank
which they employ for acquiring the parser, constitutes a rich enough
sample of the domain.  It is assumed that the part of the annotation
scheme that is actually instantiated in the tree-bank does not
under-generate on sentences of the domain.  This assumption is not met
by our current tree-bank.  It turned out that one can expect the
tree-bank grammar to generate a parse-space containing the right
syntactic/semantic tree only for approximately~90-91\% of unseen
domain utterances. This figure constitutes an upper bound on the
accuracy for any probabilistic model.  Enlarging the tree-bank does
not guarantee a good coverage, however.  The tree-bank will always
represent only a sample of the domain. A solution for this problem is
the development of automatic methods for generalising grammars, to
enhance their coverage.  The goal is to improve both accuracy and
coverage by generalising over the structures encountered in the
tree-bank.

\subsection*{Acknowledgements}
This research was carried out within the framework of the Priority
Programme Language and Speech Technology (TST). The TST-Programme is
sponsored by NWO (Dutch Organisation for Scientific Research).

\bibliographystyle{fullname}

\end{document}